\documentclass[journal]{IEEEtran}  

\usepackage{glossaries}
\usepackage{cite}
\usepackage{amsmath,amssymb,amsfonts}
\usepackage{graphicx}
\usepackage{algorithm,algorithmic}
\usepackage{hyperref}
\hypersetup{hidelinks}
\usepackage{textcomp}
\usepackage{xurl}

\glsdisablehyper

\everymath=\expandafter{\the\everymath\displaystyle}
\usepackage{times}
\usepackage{xcolor}
\usepackage{scalerel,stackengine,mathtools,bm}
\usepackage{dsfont}
\usepackage[capitalize,noabbrev]{cleveref}
\usepackage{pgfplots}
\usepackage{times}
\usepackage{siunitx}
\usepackage{xfrac}
\usepackage{caption,subcaption}
\usepackage{xparse}
\usepackage{xstring}
\usepackage{todonotes}
\usepackage{tikz}
\usetikzlibrary{positioning, arrows, calc, fit}

\newcommand{\eg}{e.\,g.}
\newcommand{\ie}{i.\,e.}
\newcommand{\diff}{\mathrm{d}}

\newcommand{\bx}{\bm{x}}
\newcommand{\hil}{\mathcal{H}}
\newcommand{\bhil}{\bm{\mathcal{H}}}
\newcommand{\x}{\bm{x}}
\newcommand{\f}{\bm{f}}
\newcommand{\y}{\bm{y}}
\newcommand{\z}{\bm{z}}
\newcommand{\e}{\bm{e}}
\newcommand{\w}{\bm{w}}
\newcommand{\h}{\bm{h}}

\newcommand{\tX}{\tilde{X}}
\newcommand{\bmu}{\bm{\mu}}
\newcommand{\bsig}{\bm{\sigma}}
\newcommand{\bvphi}{\bm{\varphi}}

\newcommand{\ts}{\textsuperscript}

\renewcommand{\footnotesize}{\scriptsize}

\DeclareMathOperator*{\tr}{tr}

\DeclareMathOperator*{\cov}{cov}

\newtheorem{theorem}{Theorem}
\newtheorem{lemma}[theorem]{Lemma}
\newtheorem{corollary}[theorem]{Corollary}
\newtheorem{assum}[theorem]{Assumption}

\newtheorem{proposition}[theorem]{Proposition}
\newtheorem{definition}[theorem]{Definition}

\newacronym{desy}{DESY}{Deutsches Elektronen-Synchrotron}
\newacronym{euxfel}{European XFEL}{European X-Ray Free-Electron Laser}
\newacronym{xfel}{XFEL}{X-Ray Free-Electron Laser}
\newacronym{lbsync}{LbSync}{laser-based optical synchronization}
\newacronym[longplural={linear matrix inequalities}]{lmi}{LMI}{linear matrix inequality}
\newacronym{lo}{LO}{local oscillator}
\newacronym{lti}{LTI}{linear time-invariant}
\newacronym{mlo}{MLO}{master laser oscillator}
\newacronym{mo}{MO}{master timing reference oscillator}
\newacronym{pi}{PI}{proportional-integral}
\newacronym{pll}{PLL}{phase-locked loop}
\newacronym{ppl}{PPL}{pump-probe laser}
\newacronym{rf}{RF}{radio-frequency}
\newacronym{rms}{RMS}{root-mean-square}
\newacronym{slo}{SLO}{subsidiary laser oscillator}
\newacronym{snr}{SNR}{signal-to-noise ratio}
\newacronym{vco}{VCO}{voltage controller oscillator}
\newacronym{inr}{INR}{improvement-to-noise ratio}
\newacronym{bo}{BO}{Bayesian optimization}
\newacronym[longplural={Gaussian processes}]{gp}{GP}{Gaussian process}
\newacronym{ei}{EI}{expected improvement}
\newacronym{rkhs}{RKHS}{reproducing kernel Hilbert space}
\newacronym{ucb}{UCB}{upper confidence bound}
\newacronym{lcb}{LCB}{lower confidence bound}
\newacronym{pdf}{PDF}{probability density function}
\newacronym{cdf}{CDF}{cumulative distribution function}
\newacronym{smgo}{SMGO-\(\delta\)}{set membership global optimization}
\newacronym{awgn}{AWGN}{additive white Gaussian noise}
\newacronym{psd}{PSD}{power spectral density}
\newacronym{lsu}{LSU}{link stabilization unit}
\newacronym{oxc}{OXC}{optical cross-correlator}
\newacronym{icm}{ICM}{intrinsic co-regionalization model}
\newacronym{lmc}{LMC}{linear model of co-regionalization}
\newacronym{mle}{MLE}{maximum likelihood estimation}
\newacronym{mcmc}{MCMC}{Markov chain Monte Carlo}
\newacronym{map}{MAP}{maximum a posteriori} 
\newacronym{lkj}{LKJ}{Lewandowski-Kurowicka-Joe}

\def\Title{An Analysis of Safety Guarantees in Multi-Task Bayesian Optimization} 
\def\Author{Lübsen}

\def\BibTeX{{\rm B\kern-.05em{\sc i\kern-.025em b}\kern-.08em
    T\kern-.1667em\lower.7ex\hbox{E}\kern-.125emX}}

\begin{document}
\title{\Title}

\author{J. O. \Author, and A. Eichler
\thanks{This work has been submitted to the IEEE for possible publication. Copyright may be transferred without notice, after which this version may no longer be accessible}
\thanks{Jannis O. Lübsen is with the Institute of Control Systems at Hamburg University of Technology, Hamburg, Germany (e-mail: jannis.luebsen@tuhh.de). }
\thanks{Annika Eichler is with the Institute of Control Systems at Hamburg University of Technology, Hamburg, Germany, and the Deutsches-Elektronen Synchrotron, Hamburg, Germany  (e-mail: annika.eichler@tuhh.de).}}

\maketitle
\begin{abstract}%
    This paper addresses the integration of additional information sources into a Bayesian optimization framework while ensuring that safety constraints are satisfied. The interdependencies between these information sources are modeled using an unknown correlation matrix. We explore how uniform error bounds must be adjusted to maintain constraint satisfaction throughout the optimization process, considering both Bayesian and frequentist statistical perspectives. This is achieved by appropriately scaling the error bounds based on a confidence interval that can be estimated from the data. Furthermore, the efficacy of the proposed approach is demonstrated through experiments on two benchmark functions and a controller parameter optimization problem. Our results highlight a significant improvement in sample efficiency, demonstrating the method’s suitability for optimizing expensive-to-evaluate functions.
\end{abstract}

In numerous practical applications, optimizing a system is often challenging due to the lack of models, often caused by the complexity of the underlying physics or the impracticality of the required identification processes.
Black-box optimization algorithms bypass the need of models for optimizations. In essence, these algorithms sequentially evaluate the black-box function for some inputs while reducing the cost. 
In the last decade, \gls{bo} has emerged as a powerful and efficient technique for tackling complex optimization problems. \gls{bo} constructs a probabilistic surrogate model to approximate an objective function with minimal assumptions. The utilization of \glspl{gp} enables the incorporation of prior knowledge about the objective function, making \gls{bo} particularly well-suited for scenarios where function evaluations are costly and observations are noisy. As a control related practical illustration of \gls{bo}, consider the optimization of a PID controller for unit step reference tracking, where the plant dynamics are unknown. A potential cost function that measures tracking accuracy could be the mean-squared error of the plant output and the step reference for a designated time window. The unknown objective is the function that maps the PID parameters to the image of the cost function. An evaluation of the objective corresponds to running the step response of the system with the specified PID parameters. The time required for a single evaluation can be in the range of seconds, depending on the bandwidth of the underlying system. The obtained cost value and the PID parameters are used to update the \gls{gp} model, which is then used to determine new promising inputs for the next evaluation.

When addressing high-dimensional optimization problems, the performance of \gls{bo} is constrained by the curse of dimensionality. This limitation manifests in the exponential growth of function evaluations required to identify the global minimum as the dimensionality of the input space increases.
Enhanced sample efficiency can be achieved by integrating low-fidelity models of the objective function into the optimization process, as demonstrated in a proof-of-concept study by \cite{Pousa2023,Letham2019}. The central tool in this approach, multi-task \gls{gp} prediction, was originally introduced by \cite{Bonilla2007}. This technique uses correlation matrices to capture the influence between various tasks, which are learned from the available data. Building on this foundation, \cite{Swersky2013} developed the first multi-task \gls{bo} algorithm. The core concept of this approach lies in incorporating auxiliary models (supplementary tasks) of the objective function, allowing the primary task to be estimated by evaluating these auxiliary models. In general, the evaluation of these auxiliary models is much less expensive in practice which can accelerate the optimization process significantly.

\begin{figure*}
    \centering
    \includegraphics[width=0.8\textwidth]{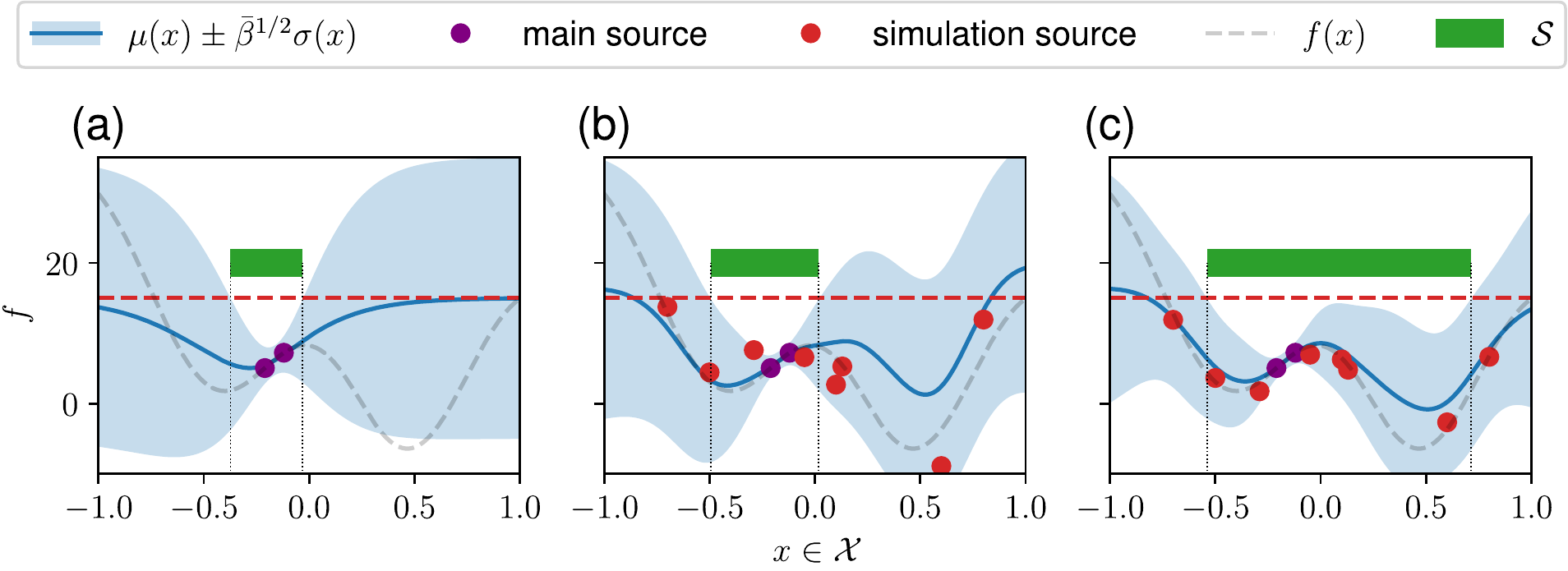}
    \caption{Overview of different safe \gls{bo} settings with safety threshold $T$ illustrated
    by "\textcolor{red}{- - -}". In (a) the single-task setting $\left(\bar{\beta} = \beta\right)$ is depicted, \ie, no simulation samples are considered,
    and the safe region is the smallest. (b) shows the multi-task setting with slight correlation and (c) with high correlation. In both cases (b) and (c), using information from an additional task increases the
    safe region.}
    \label{fig:safebo}
\end{figure*}

In many practical optimization problems, constraints must be taken into consideration to avert undesirable outcomes, such as system damage. For instance, in the PID controller illustration, it is necessary to ensure that the selected PID parameters result in a stable closed-loop system. Since the plant dynamics are unknown the constraints are unknown and must be learned online. For example, the constraint could be selected such that the function should be evaluated solely for inputs that yield function values below a predetermined threshold.
The theoretical foundation for safe \gls{bo} is rooted in the minimization of the regret via uniform error bounds in multi-armed bandit problems, as established by \cite{Snirivas2010}, and later improved in terms of performance by \cite{chowdhury17a}. The uniform error bounds are stated as 
\begin{align}
    \mathbb{P}\left\{|f(\x)-\mu(\x)|\leq \beta^{\frac{1}{2}}\sigma(\x),~\forall\,\x\in\mathcal{X}\right\}\geq 1-\delta,
    \label{eq:uniform_error_bound}
\end{align}
which means that the deviation of the unknown function $f$ can be bounded from the posterior mean $\mu$ by scaling the posterior standard deviation $\sigma$ with the factor $\beta^{\frac{1}{2}}$.
Building on the results of \cite{Snirivas2010}, \cite{Sui2015} introduced \texttt{SafeOpt}, the first method for safe \gls{bo}. A one-dimensional example of safe \gls{bo} is visualized in \cref{fig:safebo}~(a). The safe region $\mathcal{S}$ includes all inputs for which the upper bound of the confidence interval is less than the safety threshold. The algorithm is only permitted to evaluate inputs that fall with in $\mathcal{S}$. Moreover, (b) and (c) show that $\mathcal{S}$ increases if the number of evaluations increase.

The aforementioned works assume that the unknown function is deterministic and belongs to the \gls{rkhs} defined by the selected kernel $k$ which is used for the \gls{gp}. In contrast, several methodologies have been developed, such as those by \cite{Lederer2019,Sun2021}, which assume that the unknown function is a sample of a prior defined by a \gls{gp}. The former approaches align with frequentist statistics and therefore lead to a scaling factor $\beta_f$, while the latter are rooted in Bayesian statistics (for a detailed comparison, see \cite{murphy2021}), leading to a different scaling factor $\beta_b$. 
Both assumptions necessitate knowledge of the \textit{correct} kernel and hyperparameters for the derivation of their respective scaling factors \cite{fiedler2024safetysafebayesianoptimization}. In the context of uncertain hyperparameters \cite{Beckers18} examined the impact of \gls{gp} misspecifications on the prediction error under wrong prior mean and covariance functions, and \cite{fiedler21} introduced a robust approach within the frequentist framework, under the assumption that the upper bound on the \gls{rkhs}-norm remains valid.
In the Bayesian context, \cite{Capone2022} introduced a method for establishing robust bounds on hyperparameters, particularly the lengthscales of radially decreasing kernels. Through Bayesian inference, a confidence interval is derived within which the true hyperparameter lies with high probability. The uniform-error bounds can then be extended to encompass all hyperparameters within this interval. Uncertain hyperparameters are particularly critical in multi-task settings, where the predictions of the primary task is influenced by the supplementary tasks. In a prior study, \cite{Luebsen2024} demonstrated that Bayesian inference of hyperparameters can be extended to the multi-task setting, where the correlation between the true objective and the reduced models is estimated online. A key assumption in this work is that the dependency of $\beta_b$ in \eqref{eq:uniform_error_bound} on the correlation matrix is known.

The advantage of safe multi-task \gls{bo} is shown in \cref{fig:safebo}. Comparing (a) - single-task - with (b) and (c) - multi-task - we can see that $\mathcal{S}$ is increased. In (a), the algorithm requires multiple evaluations of the main task until the global minimum is contained in $\mathcal{S}$. In contrast, as illustrated in (b), a slight correlation results in an augmentation of the safe region and a reduction in the number of evaluations of the main task. For higher correlation, as depicted in (c), the safe region is further expanded. Here, the algorithm would be directly allowed to evaluate at the global minimum. 

The contributions of this manuscript are summarized as follows:
\begin{enumerate}
    \item The definitions of the scaling factors from both statistics are extended to a multi-task setting, \ie, safety guarantees using additional information are derived. It extends the work done by \cite{Luebsen2024} by deriving multi-task scaling factors which is to the best of the authors' knowledge, the first work.
    \item Improved single-task scaling factors under the Bayesian point of view are provided.
    \item Numerical comparisons between (safe) multi-task approaches and (safe) single-task approaches in the Bayesian setting are provided. The code is available on GitHub\footnote{\url{https://github.com/TUHH-ICS/2025-code-An-Analysis-of-Safety-Guarantees-in-Multi-Task-Bayesian-Optimization}}.
\end{enumerate}

\section{Fundamentals}
\label{sec:fundamentals}
In \gls{bo}, \glspl{gp} are used to model an unknown objective function \(f:\mathcal{X} \to \mathbb{R}\), where the domain \(\mathcal{X} \subset \mathbb{R}^d\) is compact. It is assumed that $f$ is continuous, which ensures an arbitrary good approximation of $f$ using universal kernels. With universal kernels we refer to kernels whose underlying \gls{rkhs} is dense in the set of continuous functions. In real applications, the function values themselves are not accessible, rather noisy observations are made. This behavior is modeled by additive Gaussian noise \(\epsilon \sim \mathcal{N}(0,\sigma_n^2)\), \ie, $y = f(\x)+\epsilon$, where $y$ is the measured value and \(\sigma_n^2\) denotes the noise variance.
Furthermore, we define the set of observations by \(\mathcal{D}\coloneqq\left\{(\x_k,y_k), k=1,\dots, N\right\}\) which is composed of the evaluated inputs combined with the corresponding observations. This set can be considered as the training set. A more compact notation of all inputs of \(\mathcal{D}\) is given by the matrix \(X = [\x_{1}, \dots, \x_{N}]^T\) and of all observations by the vector \(\y=[y_1,\dots,y_N]^T\). Using this data set, the \gls{gp} creates a probabilistic surrogate model to predict $f(\x)$. These predictions serve as inputs for an acquisition function $\alpha$, which identifies new promising inputs likely to minimize the objective.
Some common choices for acquisition functions are \gls{ucb}, (log) expected improvement \cite{Jones1998}, or predictive entropy search \cite{hernandez14}. 

\subsection{Gaussian Processes}
\label{subsec:gaussian_processes}

A \gls{gp} is fully defined by a mean function \(m(\x)\) and a kernel \(k(\x,\x'): \mathcal{X}\times\mathcal{X} \to \mathbb{R}\). In the context of \glspl{gp}, the kernel is also referred to as a covariance function. The difference between those two concepts lies in the fact that a covariance function is positive definite, whereas a kernel is not necessarily so. However, throughout this work, the term \textit{kernel} is used to refer to a positive definite kernel. Positive definiteness in terms of kernels means that the resulting Gram matrix $[K]_i,j = k(\x_i,\x_j)$ is positive semidefinite for all $\x_i,\x_i\in\mathcal{X}$ \cite{Schoelkopf2002}. The prior distribution of the function values $f(\x)$ are given by $p(f(\x)) = \mathcal{GP}(0,k(\x,\x'))$ where $m(\x)$ is replaced by the zero function without loss of generality. The kernel determines the dependency between function values at different inputs which is expressed by the covariance operator 
\begin{align}
\mathrm{cov}(f(\x),f(\x'))=k(\x,\x').
\label{eq:single_task_cov}
\end{align} 
Commonly used kernels are the spectral mixture \cite{wilson13}, Mat\'ern \cite{Genton01} or squared exponential kernel, where the latter is defined as $k_\mathrm{SE}(\x,\x') = \sigma_f^2\exp\left(-\frac{1}{2}(\x-\x')^T\Delta^{-2}(\x-\x')\right)$ with \(\Delta = \mathrm{diag}(\bm{\vartheta}) = \mathrm{diag}([\vartheta_1,\dots,\vartheta_d]^T)\).
The signal variance \(\sigma_f^2\), the lengthscales \(\bm{\vartheta}\) and the noise variance \(\sigma_n^2\) constitute the hyperparameters, allowing for adjustments of the kernel.

Given the set of observations and the prior of $f(\x)$, the posterior $p(f(\x)|X,\y) = \mathcal{N}(\mu(\x),\sigma^2(\x))$ can be computed by applying Bayes' rule.
As shown by \cite{Williams2006} the posterior is also Gaussian given by
\begin{equation}
\label{eq:posterior_gp}
\begin{aligned}
        \mu(\x) &= K_X(\x)\left(K+\sigma_n^2I\right)^{-1}\y\\
        \sigma^2(\x) &= k(\x,\x)-K_X(\x)\left(K+\sigma_n^2I\right)^{-1}K_X(\x)^T,   
\end{aligned}
\end{equation}     
where $K_X(\x) = k(\x,X)$ and $K = k(X,X)$ is the Gram matrix of the training data.

An examination of \eqref{eq:posterior_gp} reveals that the kernel $k(\x,\x')$ plays a crucial role in the field of kernel regression in general. The kernel implicitly maps the inputs into a high-dimensional feature space, the \gls{rkhs}, and computes an inner product very efficiently thanks to the kernel trick. More concretely, it can be shown that all positive semidefinite kernels can be represented by 
\begin{align}
     k(\x,\x') = \langle \bvphi(\x),\bvphi(\x')\rangle = \sum_{i=1}^{N_\mathcal{H}} \lambda_i\psi_i(\x)\psi_i(\x),
     \label{eq:kernel_expansion}
 \end{align} 
where $\bvphi(\x):\mathcal{X}\rightarrow \ell_2^{N_\mathcal{H}} := (\sqrt{\lambda_i}\psi_i(\x))_{i\in 1,\dots,N_\mathcal{H}}$ and $N_\mathcal{H}$ is the length of the sequence which can be infinite. We use the symbol $\mathcal{H}$ to denote Hilbert spaces. The given formalism follows directly from Mercers Theorem \cite{Mercer1909}. In summary, it states that a positive semidefinite kernel can be expressed in terms of a dot product in $\ell_2^{N_\mathcal{H}}$ which is a space of sequences of eigenfunctions $\sqrt{\lambda_i}\psi_i(\x) \in L_2(\mathcal{X})$. Since the goal is to construct a \gls{rkhs}, the evaluation functional need to be bounded and represented as an inner product. The former is clearly fulfilled by \eqref{eq:kernel_expansion} and the latter is defined by 
\begin{align}
    \label{eq:evaluation_functional}
    f(\x) = \langle\f,\bvphi(\x)\rangle_\hil.
\end{align} 
The eigenfunctions are also members of the \gls{rkhs}, which implies $\psi_i(\x)=\langle\sqrt{\lambda_i} \psi_i,\bvphi(\x)\rangle_\hil$ and the expansion coefficients $\psi_i$ are chosen to be orthogonal with respect to the inner product, \ie, $\langle \sqrt{\lambda_i}\psi_i,\sqrt{\lambda_j}\psi_j\rangle  = \rho_{ij}$. Thus, the dot product that fulfills this requirement is defined as $\langle \psi_i,\psi_j\rangle = \rho_{ij}/\lambda_i$. $\psi_i$ can be considered to be an $N_\hil$ dimensional vector with a unique $1$ at the $i$\ts{th} entry and zero otherwise. Moreover, it can be shown that each \gls{rkhs} corresponds uniquely to a kernel according to the Moore-Aronszajn Theorem \cite{aronszajn50reproducing}. Using the eigenfunction expansion coefficients which form an orthonormal basis, all elements contained in the Hilbert space can be written as $\f = \sum_{i=1}^{N_{\mathcal{H}}}\alpha_i\sqrt{\lambda_i}\psi_i$, where $\alpha_i\in\mathbb{R}$. 
It should be noted, however, that depending on the initial assumptions (Bayesian or frequentist), $\f$ does not necessarily lie in the \gls{rkhs} if $N_\mathcal{H}\rightarrow \infty$ as shown by \cite{Williams2006}. We will also elaborate on this in \cref{subsec:b_vs_f}.

\subsection{Multi-Task Gaussian Processes}
From now, vector valued functions $\f = [f_1,\dots, f_u]: \mathcal{X} \to \mathbb{R}^u$ are considered where each output entry represents the function from a different task, $f_1$ denotes the primary task and $f_2,\dots,f_u$ the supplementary tasks.
To tackle inter-function correlations, the kernel from \eqref{eq:single_task_cov} is extended by additional task inputs, \ie, $k((\x,z),(\x',z'))=\mathrm{cov}(f_z(\x),f_{z'}(\x'))$ where $z,z'\in\{1,\dots,u\}$ denote the task indices and $u$ is the total number of tasks. If the kernel can be separated into a task-depending $k_z$ and an input-depending kernel $k_x$, \ie, $k((\x,z),(\x',z')) = k_z(z,z')\, k_x(\x,\x')$, then the kernel is called separable which is used in most literature, e.g., \cite{Bonilla2007,Letham2019,Swersky2013}. If there is a single base kernel $k_x$, this is denoted as the \gls{icm} \cite{Alvarez2012}.

More generally, separable covariance functions can be described as 
\begin{equation}
\begin{alignedat}{4}
K: &\mathcal{X} \times \mathcal{X} &&\to && \mathcal{L}_+(\mathbb{R}^u)\\  
    &\x,\x' &&\mapsto && \Sigma k(\x,\x'),
\end{alignedat}
\end{equation}

where $k$ denotes a scalar kernel, $\mathcal{L}_+(\mathbb{R}^u) := \mathcal{L}_+(\mathbb{R}^u,\mathbb{R}^u): \mathbb{R}^u\to\mathbb{R}^u$ is the set of positive definite linear operators on the vector field $\mathbb{R}^u$, and $\Sigma \in \mathcal{L}_+(\mathbb{R}^u)$, $\Sigma = \Sigma^{\frac{1}{2}}\Sigma^{\frac{1}{2}}$ is a symmetric positive definite matrix. Furthermore, we introduce the tensor vector space $\mathcal{W}\bigotimes \mathcal{H}$ in accordance to \cite{Caponnetto2008} spanned by elements of the form $\w\otimes \h$ with $\w\in\mathcal{W}\subseteq \mathbb{R}^u$ and $\h\in\mathcal{H}$. All elements $\w_1\otimes \h_1,\w_2\otimes \h_2\in\mathcal{W}\bigotimes \mathcal{H}$ and constant $c\in\mathbb{R}$ satisfy the multilinear relation
 \begin{align*}
     c(\w_1+\w_2)\otimes \h_1 = c(\w_1\otimes \h_1)+c(\w_2\otimes \h_1)
 \end{align*}
 and vise versa. To obtain a dot product space, we define 
 \begin{align*}
     \langle \w_1\otimes \h_2,\w_2\otimes \h_2\rangle = \langle \w_1,\w_2\rangle_\mathcal{W}\langle \h_1,\h_2\rangle_\mathcal{H},
 \end{align*}
 and take the completion with respect to the induced norm of the elements in $\mathcal{W}\bigotimes \mathcal{H}$. Thus, the tensor vector space $\mathcal{W}\bigotimes \mathcal{H}$ becomes a Hilbert space $\bhil$.

 In the next step, we show that $\bhil$ indeed defines a \gls{rkhs}. Therefore, similar to the single-task case we define the feature map
 \begin{equation}
    \label{eq:feature_map_MT}
    \begin{alignedat}{3}
        \Phi(\bx):& \mathcal{X} & \to &\mathcal{L}(\mathcal{W},\bhil)\\
        \Phi(\bx)\w: & \x &\mapsto &\Sigma^{\frac{1}{2}}\w\otimes \bvphi(\x) 
    \end{alignedat}
 \end{equation}
for all $\w\in\mathcal{W}$, and the adjoint feature map  
\begin{equation}
    \label{eq:adjoint_feature_map_MT}
    \begin{alignedat}{3}
       \Phi^*(\bx): & \mathcal{X} & \to &\mathcal{L}(\bhil,\mathcal{W})\\
        \Phi^*(\bx) \w\otimes \h: & \x & \mapsto &\Sigma^{\frac{1}{2}} \w\langle\bvphi(\x),\h\rangle_\mathcal{H}
    \end{alignedat}
\end{equation}
for all $(\w\otimes \h)\in\bhil$, where $\mathcal{L}(\mathcal{W},\bhil)$ denotes the set of linear operators from $\mathcal{W}\to\bhil$ and $\mathcal{L}(\bhil,\mathcal{W})$ is defined analogously. With the two definitions it is to check that $\Phi$ induces a multi-task kernel $K$. This follows by \eqref{eq:feature_map_MT} and \eqref{eq:adjoint_feature_map_MT}
 \begin{alignat*}{3}
     \langle\Phi(\x),\Phi(\x')\rangle_{\bhil_\Sigma} &= \Phi^*(\x)\Phi(\x')
      &&=  \Sigma^{\frac{1}{2}} \Sigma^{\frac{1}{2}}\langle \bvphi(\x),\bvphi(\x')\rangle_\mathcal{H}\\
       &= \Sigma k(\x,\x')
       &&= K_\Sigma(\x,\x').
 \end{alignat*}
The evaluation functional can be represented by the dot product
\begin{equation}
 \begin{aligned}
    \f(\x) = \langle\Phi(\x),\f_w\otimes \f_h\rangle_{\bhil_\Sigma} &= \Sigma^{\frac{1}{2}}\f_w\langle\bvphi(\x),\f_h\rangle_\mathcal{H}\\ &= \Sigma^{\frac{1}{2}}\f_w f(\x). 
 \end{aligned}
\end{equation}

 Since, we will deal with multiple different covariance matrices $\Sigma$, we use the notation $\langle\cdot,\cdot\rangle_{\bhil_\Sigma}$ to denote in which \gls{rkhs} the inner product is taken.

Moreover, it is assumed that for each task there exists a data set $\mathcal{D}_i$ which are stacked into a global set $\bm{\mathcal{D}} \coloneqq \{\tX,\tilde{\y},Z\}$, where $\tX = [X_1^T,\dots,X_u^T]^T$, $\tilde{\y} = [\y_1^T,\dots,\y_u^T]^T$ and $Z=[\e_1,\dots,\e_2,\dots,\e_u]$. $\e_i$ denote the canonical basis vectors of $\mathbb{R}^u$ and act as the task indices. The Gram matrix is given by
\begin{align}
\label{eq:multi_task_gram_matrix}
\bm{K} = K(\tX,\tX)= 
\begin{bmatrix} 
\Sigma_{1,1}^2K_{1,1} & \dots & \Sigma_{1,u}^2K_{1,u}\\
\vdots & \ddots & \vdots \\
\Sigma_{u,1}^2K_{u,1} & \dots & \Sigma_{u,u}^2K_{u,u}
\end{bmatrix},    
\end{align}
where $K_{\z,\z'}$ are Gram matrices using data from tasks $\z$ and $\z'$. According to \eqref{eq:kernel_expansion}, the Gram matrix can be expressed as $\bm{K} = K_\Sigma(\tX,\tX) =  \Phi_\Sigma(\tX)^*\Phi_\Sigma(\tX)$, where \begin{align*}\Phi_\Sigma(\tX) = (\Sigma^{\frac{1}{2}}\z_i \otimes \bvphi(\x_i))_{i=1}^n.
\end{align*}
We use $\Phi_{\Sigma,\tilde{X}}=\Phi_\Sigma(\tX)$ or just $\Phi_{\tX}$ if $\Sigma$ is clear from the context.
Note that if the covariance entries are zero, \ie, $\Sigma_{\z,\z'} = 0, \forall \z \neq \z'$, the off-diagonal blocks of $\bm{K}$ are zero which means that all tasks are independent and can be divided into separate \glspl{gp}. To perform inference, one simply needs to substitute the single-task Gram matrix $K$ and measurements $\y$ in \eqref{eq:posterior_gp} by their multi-task equivalents $\bm{K}_\Sigma$ and $\tilde{\y}$. 

This section draws a connection between the single-task and multi-task setting. The feature map lifts the inputs to a tensor vector space which obeys a \gls{rkhs}. Throughout this manuscript, the inner product of $\mathcal{W}$ corresponds to the usual dot product.

\section{Safe Bayesian Optimization}
\label{sec:safe_bo}
In this section, we will review the concept of safe \gls{bo} under the Bayesian and frequentist statistics. In this context safeness is defined as in \eqref{eq:uniform_error_bound}.
The goal is to define a confidence $\beta^\frac{1}{2}$ which includes the deviation of the unknown function from the posterior mean \eqref{eq:posterior_gp} with probability $1-\delta$. First, we start in \cref{subsec:b_vs_f} with a short explanation of the assumptions required and highlight their differences. Then, we switch to safeness and start with a short explanation of the derivation of the uniform error bounds for the single-task case and then extend the results to the multi-task setting. This is a requisite step as the scaling factor $\beta$ for the single-task case also depends on the correlation matrices when considering multiple tasks. We begin with the frequentists perspective in \cref{subsec:frequentist_statistics} and then proceed to the Bayesian in \cref{subsec:bayesian_statistics}. Throughout the whole manuscript, we restrict the space of $\Sigma$ to positive correlation matrices $\mathcal{C}:=\{\Sigma\in\mathcal{L}_+(\mathcal{W})|\Sigma_{i,j} \geq 0, ~\forall i,j\}$.

\subsection{Bayesian vs Frequentist}
\label{subsec:b_vs_f}
In frequentist statistics, it is assumed that the unknown function $\f\in\hil$ is deterministic and a member of the \gls{rkhs} of the kernel $k$. As we have seen in \cref{sec:fundamentals} the evaluation of the function $\f$ at point $\x$ can be written as $f(\x) = \langle \f, \bvphi(\x)\rangle_{\hil}$. Since a \gls{rkhs} is complete the \gls{rkhs}-norm $\|\f\|_{\hil}^2 = \langle \f,\f\rangle_\hil = \sum_i f_i^2 < \infty$ is bounded for all its members. The determinism of $\f$ implies that the randomness is not induced from the model but from the data. This is contrary to the Bayesian setting, where also the model is random. In particular, in Bayesian statistics the function values have a prior given by 
\begin{align}
    \label{eq:sample_of_prior}
    f(\x) \sim \mathcal{GP}(0,k(\x,\x')).
\end{align} 
Considering the evaluation functional in \eqref{eq:evaluation_functional}, \eqref{eq:sample_of_prior} implies that the coefficient vector $\f$ is not deterministic but stochastic. This can be shown by assuming the coefficient entries to be i.i.d with a standard Gaussian distribution with zero mean and unit variance, in other words, $f_i\sim\mathcal{N}(0,1)$ for all $i$ and $\cov(f_i,f_j) = 0$ for all $i\neq j$. Since, $f(\x)$ is linear in the coefficients, we know that the prior of $f(\x)$ is also Gaussian. Obviously, the mean remains zero and the variance is given by $\mathbb{V}[f(\x)] = \mathbb{E}[f(\x)^2]=\mathbb{E}[\bvphi(\x)^T\f\f^T\bvphi(\x)] = \mathbb{E}[\tr(\f\f^T\bvphi(\x))\bvphi(\x)^T] = \tr(I\bvphi(\x)\bvphi(\x)^T) = k(\x,\x')$. Comparing this to \eqref{eq:sample_of_prior}, we observe that we recovered the aforementioned prior of $f(\x)$ given by the \gls{gp}. From this, we see that \glspl{gp} naturally operate in the Bayesian setting, where the prior, as given in equation \eqref{eq:sample_of_prior}, and the likelihood together enable the computation of the posterior \eqref{eq:posterior_gp}.
Another interesting characteristic occurs if the expected norm of the stochastic coefficient vector is investigated. More concretely, we have 
\begin{align}
    \mathbb{E}[\langle\f,\f\rangle] = \sum_{i=1}^{N_\hil} \mathbb{E}[f_i^2] = \sum_{i=1}^{N_\hil} 1.
    \label{eq:norm_of_sample}
\end{align}
The series in \eqref{eq:norm_of_sample} becomes infinity if the dimension of $\f$ is infinite. This implies that samples from the \gls{gp} do not belong to the \gls{rkhs} almost surely if the \gls{rkhs} is infinite-dimensional \cite{Williams2006}.
Moreover, this means that the set of considered functions is larger than in the frequentist case. Consequently, the posterior of a \gls{gp} does not offer predictions for function values belonging to functions from a \gls{rkhs}, but rather, are samples from a \gls{gp} as in $\eqref{eq:sample_of_prior}$.

We have seen that the main difference between the two statistics lies in the assumption about the model, which is deterministic in the frequentist and random in the Bayesian setting. Clearly, the use of a particular statistic implies assumptions about the underlying true system.

\subsection{Frequentist Statistics}
\label{subsec:frequentist_statistics}
\subsubsection*{Single-Task}
In the frequentist literature of safe \gls{bo} the definition of the uniform error bound is slightly changed, i.e., 
\begin{align*}
    \mathbb{P}\{|f(\x)-\mu_t(\x)| \leq \beta_{f,t}^\frac{1}{2}\sigma_t(\x),\quad \forall \x\in\mathcal{X},\,\forall t\in\mathbb{N}\}\leq 1-\delta,
\end{align*}
where the scaling factor $\beta_{f,t}$ must also hold for all iterations $t\in\mathbb{N}$. 
A detailed derivation of $\beta_{f,t}$ can be found in \cite{chowdhury17a,Snirivas2010,fiedler21}. Throughout this manuscript we are not interested in guarantees for all iterations but only for the current one. Accordingly, the construction of a martingale sequence as done by \cite{chowdhury17a} is replaced by a concentration inequality \cite{Hsu2011}. Nevertheless, the following results presented in this work can be easily extended to apply for all iterations. For the ease of the notation, we omit the explicit dependence of the posterior mean and variance on the iteration $t$ and write $\mu(\x)$, $\sigma^2(\x)$ instead. In accordance with the previous discussion, we state the following assumption.
\begin{assum}
\label{assum:frequentist_st}
The unknown function $\f$ lives in a reproducing kernel Hilbert space $\hil$ defined by the selected kernel $k$. In addition, the norm of $\f$ is known, \ie, $\|\f\|_{\hil} = M$.     
\end{assum}
The first part of this assumption can be easily satisfied by choosing a kernel that fulfills the universal approximation property. The second part is hard to satisfy in practice, because the function is unknown and hence its norm in the \gls{rkhs} \cite{fiedler2024safetysafebayesianoptimization}. Usually $M$ is guessed to upper bound the true norm which introduces additional conservatism.
To avoid redundancy in the derivation of the uniform-error bounds, we proceed directly with the multi-task setting.

\subsubsection*{Multi-Task}
For the multi-task case, we extend the assumptions made in Assumption~\ref{assum:frequentist_st} by the following.
\begin{assum}
    \label{assum:frequentist_mt}
    The unknown function $\f$ lives in a reproducing kernel Hilbert space $\bhil$ defined by the selected kernel $K$. In addition, $\f$ is a linear combination of $u$ latent functions $\h_i\in\hil\, , \forall i$ and known norms $\|\h_i\|_\hil$.     
\end{assum}
To avoid any further restrictiveness, we assume that the norms of the latent functions are known which corresponds to the same degree of knowledge that is required as in Assumption~\ref{assum:frequentist_st}. Alternatively, it may be a valid assumption that the norm of $\f$ is known for some $\Sigma$.\\
For vector valued functions we define the uniform error bounds as
\begin{align*}
    |\f(\x)-\bmu^{\Sigma'}(\x)| \leq \beta^\frac{1}{2}_f(\Sigma')\bsig^{\Sigma'}(\x),
\end{align*}
where $\bmu^{\Sigma'}(\x) = [\mu_{1}(\x),\dots,\mu_{u}(\x)]^T$ and $\bsig^{\Sigma'}(\x) = [\sigma_{1}(\x),\dots,\sigma_{u}(\x)]^T$. 
The superscript $\Sigma'$ in, \eg, $\bmu^{\Sigma'}(\x)$ emphasizes that a multi-task kernel $K(\x,\x')=\Sigma' k(\x,\x')$ with correlation matrix $\Sigma'$ is used for inference of the posterior. 
It is important to note that $\bmu^{\Sigma'}(\x)$ and $\bmu^{\Sigma}(\x)$ are different functions.

In the first, we derive multi-task uniform error bound for a given correlation matrix $\Sigma'$.
\begin{lemma}
    \label{lemma:frequentist_multi-task}
    Let $\f$ be a vector valued function that lives in $\bhil_{\Sigma'}$ which is the \gls{rkhs} with a correlation factor $\Sigma' = \Sigma'^{\frac{1}{2}}\Sigma'^{\frac{1}{2}}$. Then, the scaling factor $\beta_f(\Sigma')$ for the multi-task setting is given by
    \begin{align}
        \label{eq:beta_frequentist}
        \beta_f(\Sigma') =\left(\|\f_{\Sigma'}\|_{\bhil_{\Sigma'}}+\sqrt{N+2\alpha\sqrt{N}+2\alpha^2}\right)^2,
    \end{align}
    where $\alpha = \sqrt{\ln{1/\delta}}$, $\|\f_{\Sigma'}\|_{\bhil_{\Sigma'}}$ is the \gls{rkhs}-norm of $\f_{\Sigma'}$ in $\bhil_{\Sigma'}$, and $\delta$ is the failure probability.
\end{lemma}
\begin{IEEEproof}
    The proof is given in \cref{proof:lemma_frequentist_multi-task}
\end{IEEEproof}

Observe that the term of $\beta_f(\Sigma')$ includes the norm $\|\f_{\Sigma'}\|_{\bhil_{\Sigma'}}$ which is not known according to Assumption~\ref{assum:frequentist_mt}.
\subsubsection*{Bound $\|\f_{\Sigma'}\|$}
The $\Sigma'$ in the subscript of $\f_{\Sigma'}$ denotes that $\f_{\Sigma'}$ is the expansion vector of $\f(x)$ in $\bhil_{\Sigma'}$, \ie, $\f(\x) = \langle \Phi(\x),\f_{\Sigma'}\rangle_{\bhil_{\Sigma'}}$. 
According to Corollary~\ref{corollary:equivalent_RKHS} we know that $\bhil_{\Sigma'} = \bhil_{\Sigma'}$ for all $\Sigma,\Sigma'$, which means that all Hilbert spaces are equivalent. 
This implies that for $\f(\x)$ there are different expansions vectors in each \gls{rkhs} such that $\f(\x) = \langle \Phi(\x),\f_{\Sigma'}\rangle_{\bhil_{\Sigma'}} = \langle \Phi(\x),\f_{\Sigma}\rangle_{\bhil_{\Sigma}}$. 
To compare norms in different \glspl{rkhs}, we define a linear operator $L:\bhil_{\Sigma}\to\bhil_{\Sigma'}$ with the following properties. 
\begin{lemma}
    \label{lemma:linear_oper} 
    The linear operator that satisfies $\langle \Phi(\x),\f_{\Sigma}\rangle_{\bhil_{\Sigma}} = \langle \Phi(\x),L\f_{\Sigma}\rangle_{\bhil_{\Sigma'}}$ has the properties
    \begin{align*}
        \bm{(i)}&\quad L\f_{\Sigma} = (\Sigma'^{-\frac{1}{2}}\Sigma^{\frac{1}{2}})\f_w\otimes \f_h,\\
        \bm{(ii)}&\quad \|L\| = \sqrt{\|\Sigma'^{-1}\Sigma\|_2}\\
    \end{align*}
\end{lemma}
\begin{IEEEproof}
    $\bm{(i)}$ Due to the equivalence of the \glspl{rkhs} we have
    \begin{alignat*}{2}
        &\langle \Phi(\x),\f_{\Sigma'}\rangle_{\bhil_{\Sigma'}} &&= \langle \Sigma'^{\frac{1}{2}}\otimes \bvphi(\x),\f'_w\otimes \f_h\rangle_{\bhil_{\Sigma'}} \\
        &\langle \Phi(\x),\f_{\Sigma}\rangle_{\bhil_{\Sigma}} &&= \langle \Sigma^{\frac{1}{2}}\otimes \bvphi(\x),\f_w\otimes \f_h\rangle_{\bhil_{\Sigma}}.
    \end{alignat*} 
    Hence, it is easy to see that $\Sigma'^{\frac{1}{2}}\f'_w = \Sigma^{\frac{1}{2}}\f_w$ which implies $\f'_w = \Sigma'^{-\frac{1}{2}}\Sigma^{\frac{1}{2}}\f_w$. 

    $\bm{(ii)}$ The second step is to derive the operator norm of $L$.
    \begin{align*}
       \|L\| &= \sup_{\|\f_{\Sigma}\|_{\bhil_{\Sigma}}=1}\|L\f_{\Sigma}\|_{\bhil_{\Sigma'}}\\ &= \sup_{\|\f_\Sigma\|_{\bhil_{\Sigma}}=1}\|(\Sigma'^{-\frac{1}{2}}\Sigma^{\frac{1}{2}})\f_w\otimes \f_h\|_{\bhil_{\Sigma}}\\
        &= \sup_{\|\f_\Sigma\|_{\bhil_{\Sigma}}=1}\|(\Sigma'^{-\frac{1}{2}}\Sigma^{\frac{1}{2}})\f_w\|_2\|\f_h\|_{\hil}\\ &\leq \|(\Sigma'^{-\frac{1}{2}}\Sigma^{\frac{1}{2}})\|_2\\
        &= \sqrt{\|(\Sigma'^{-\frac{1}{2}}\Sigma^{\frac{1}{2}})^T(\Sigma'^{-\frac{1}{2}}\Sigma^{\frac{1}{2}})\|_2}\\
        & = \sqrt{\|\Sigma'^{-1}\Sigma\|_2}.
    \end{align*}

    
\end{IEEEproof}

Lemma~\ref{lemma:linear_oper} allows representing $\f$ in different \gls{rkhs}. Now, this result can be used to bound the \gls{rkhs}-norm for any $\Sigma$ as summarized in the following lemma. 
\begin{lemma}
    \label{lemma:bound_f}
    Let $\f$ be a vector valued function and let $\bhil_\Sigma, \bhil_{\Sigma'}$ be the natives spaces of the kernels $K_\Sigma$ and $K_{\Sigma'}$, respectively. Then, with $\lambda = \sqrt{\|\Sigma'^{-1}\Sigma\|_2}$ we have
    \begin{align}
        \label{eq:inequality_of_rkhs_norms}
        \lambda\|\f_{\Sigma}\|_{\bhil_{\Sigma}} \geq \|\f_{\Sigma'}\|_{\bhil_{\Sigma'}}
    \end{align}
\end{lemma}
\begin{IEEEproof}
    \begin{align*}
       \|\f_{\Sigma'}\|_{\bhil_{\Sigma'}} = \|L\f_{\Sigma}\|_{\bhil_{\Sigma'}} \leq \|L\|\|\f_{\Sigma}\|_{\bhil_{\Sigma}}.
    \end{align*}
    The result follows from Lemma~\ref{lemma:linear_oper}.
\end{IEEEproof}

Lemma~\ref{lemma:bound_f} shows how the \gls{rkhs}-norm of $\f$ can be bounded if the norm for some correlation matrix $\Sigma$ is known. Recalling Assumption~\ref{assum:frequentist_mt} and the norm in Lemma~\ref{lemma:frequentist_multi-task}, it is clear that the resulting scaling factor to bound $\|f_{\Sigma'}\|_{\bhil_{\Sigma'}}$ in terms of $\|\h_i\|_\hil$ is a special case of Lemma~\ref{lemma:bound_f}, where $\Sigma$ is the identity matrix. For convenience, this is summarized in the following proposition. 
\begin{proposition}
    \label{prop:frequentist_uniform_error_bound}
     Let $\f$ be a vector valued function which lives in the Hilbert spaces $\bhil_\Sigma,\bhil_{\Sigma'}$. Then with 
     \begin{align*}
        &\bar{\beta}_{f} = \left(\lambda\|\h\|_{\bhil}+\sqrt{N+2N\sqrt{\ln\frac{1}{\delta}}+2\ln\frac{1}{\delta}}\right)^2,
    \end{align*}
        where $\lambda = \sqrt{\|\Sigma'^{-1}\|}$, we have
     \begin{align*}
          |\f(\x)-\bmu^{\Sigma'}(\x)| \leq \bar{\beta}_{f}^\frac{1}{2}\bsig^{\Sigma'}(\x),~ \forall x\in\mathcal{X}
      \end{align*} 
      with probability $1-\delta$.
 \end{proposition} 
 \begin{IEEEproof}
     Setting $\Sigma^{\frac{1}{2}}$ to identity and using Lemma~\ref{lemma:bound_f} together with Lemma~\ref{lemma:frequentist_multi-task} gives the result. 
 \end{IEEEproof}
Proposition~\ref{prop:frequentist_uniform_error_bound} allows to bound the deviation of the true function from the posterior mean for any correlation matrix that is used during inference. Alternatively, if all inner products of the latent functions are known, \ie, $\langle \h_i,\h_j\rangle_\hil$, the norm can be calculated exactly \cite{Alvarez2012} to
\begin{align*}
    \|\f_\Sigma\|_{\bhil_\Sigma} = \sum_{i,j=1}^{u}\Sigma_{i,j}^{-1}\langle \h_i,\h_j\rangle_\hil,
\end{align*}
where $\Sigma_{i,j}^{-1}$ is the $(i,j)$ entry of $\Sigma^{-1}$. In the frequentist setting, any correlation matrix $\Sigma'$ can be used for inference as long as the \gls{rkhs}-norm is scaled appropriately.



\subsection{Bayesian Statistics}
\label{subsec:bayesian_statistics}
In this subsection the safeness problem is view from a Bayesian point of view, \ie, we assume that the unknown function $f$ is a sample of a \gls{gp} with a prior given by \eqref{eq:sample_of_prior}. The section starts with an improvement of the derivation of the single-task scaling factor $\beta_b$ from \cite{Lederer2019}. Afterwards, the multi-task case is considered and main result of manuscript is stated. 

\subsubsection*{Single-Task}
There exist multiple approaches for obtaining the scaling factor in the Bayesian setting. The work of \cite{Snirivas2010} is highlighted for its derivation of frequentist bounds, while also Bayesian bounds are considered. Another approach, presented by \cite{Lederer2019} involves the discretization of the compact vector space $(\mathcal{X},\|\cdot\|_p)$ with $p \in \mathbb{N}$ into a finite set $\mathcal{I}$. This can be considered as a set of equivalence classes where each member $[x]$ is defined as $[x]:=\{a\in\mathcal{X}| \|x-a\|_p\leq \tau\}$. Since $\mathcal{X}$ is compact and $\tau >0$ this leads to a quotient set with finite cardinality $|\mathcal{I}| < \infty$. Then, on this finite set, a concentration inequality, \eg, Chernoff bound \cite{Chernoff1952}, can be applied to obtain a confidence region wherein $f$ lies with high probability for all $x\in\mathcal{I}$. The discretization error between the equivalence classes is addressed by using Lipschitz continuity of the posterior mean/variance and sample function. In order to ensure the latter, we make the following assumption.
 \begin{assum}
        \label{assum:continuity}
        The unknown function $f(\x): \mathcal{X}\to \mathbb{R}$ is a sample of a \gls{gp} with kernel $k(\x,\x')$ which has at least four partial derivatives on the compact vector space $(\mathcal{X},\|\cdot\|_p)$.  
 \end{assum}
 The first part of the assumption shows the Bayesian setting, \ie, the unknown function is stochastic. The second part ensures that samples from the \gls{gp} are Lipschitz continuous \cite{Ghosal2006}.

The previous argumentation relies heavily on the Lipschitz continuity of posterior distribution and the kernel. Hence, for clarity, the definition of those quantities can be read as follows:
\begin{definition}
    \label{definition:Lipschitz_constant_kernel}
    The Lipschitz constant of the kernel $k(\x,\x')$ is defined as
    \begin{align*}
        |k(\x,\x') - k(\y,\x')| \leq L_k\|\x-\y\|_p,~\forall \x,\x',\y\in\mathcal{X}.
    \end{align*}
    In addition, the modulus of continuity for the posterior mean is defined as
    \begin{align*}
        |\mu(\x)-\mu(\y)| \leq \omega_{\mu}(\|\x-\y\|_p),
    \end{align*}
    analogously for the posterior standard deviation $\sigma(\x)$.
\end{definition}

\begin{proposition}
    \label{proposition:Multi-task_uniform_error_bound}
   Let Assumption~\ref{assum:continuity} hold and $L_k$ as in Definition~\ref{definition:Lipschitz_constant_kernel}. By continuity of $k(\x,\x')$ the posterior mean function $\mu$ and standard deviation 
    $\sigma$ of a \gls{gp} conditioned on the training data $\mathcal{D}:=\{X,\y\}$ are continuous with moduli of continuity $\omega_{\mu}$ and $\omega_{\sigma} $ on $\mathcal{X}$, respectively, given by
    \begin{align*}
    \bm{(i)}\quad \omega_{\mu}(\tau) &\leq \sqrt{2\tau L_k}\|\bm{\mu}\|_\hil\\
    \bm{(ii)}\quad \omega_{\sigma}(\tau) &\leq \sqrt{2\tau L_k}.
    \end{align*}
    Moreover, choose $\delta\in (0,1)$, $\tau\in\mathbb{R}_+$ to obtain $\mathcal{I}$, and set 
    \begin{align*}
    \beta_b&=2\log\left(\frac{|\mathcal{I}|}{\delta}\right)\\
    \gamma(\tau)&=L_f\tau+\omega_\mu(\tau)+\beta_b^\frac{1}{2}\omega_{\sigma}(\tau),
    \end{align*}
    where $L_f$ is the Lipschitz constant of the unknown function $f$.
    Then, it holds that
    \begin{align*}
    \bm{(iii)}\quad \mathbb{P}&\left\{|f(\x)-\mu(\x)|\leq 
    \beta_b^\frac{1}{2}\sigma(\x)+\gamma(\tau), 
    ~\forall\x\in\mathcal{X}\right\}\\&\geq 1-\delta.
    \end{align*}
\end{proposition}

\begin{IEEEproof}
    $\bm{(i)}$ Since the posterior mean lies in the \gls{rkhs} of $k(\x,\x')$, by the reproducing property 
    \begin{alignat}{5}
        \label{eq:mean_lipschitz_derivation}
        \begin{split}
            &|\mu(\x)-\mu(\x')| = |\langle\bvphi(\x),\bm{\mu}\rangle - \langle\bvphi(\x'),\bm{\mu}\rangle|\\
            &\quad= |\langle\bvphi(\x)-\bvphi(\x'),\bm{\mu}\rangle|\\
             &\quad\leq \sqrt{\langle\bvphi(\x)-\bvphi(\x'),\bvphi(\x)-\bvphi(\x')\rangle} \|\bm{\mu}\|_\hil\\
            &\quad = \sqrt{k(\x,\x)-2k(\x',\x) + k(\x',\x')} \|\bm{\mu}\|_\hil\\
            &\quad\leq \sqrt{2L_k\|\x-\x'\|_p}\|\bm{\mu}\|_\hil.
        \end{split}
    \end{alignat}
    
    $\bm{(ii)}$ To determine $\omega_\sigma$, we can apply a similar reasoning. Firstly, note that $|\sigma^2(\x)-\sigma^2(\x')|\geq |\sigma(\x)-\sigma(\x')|^2$, which can be derived from the non-negativity of the standard deviation. Additionally, we can establish that the posterior variance also lies within the \gls{rkhs} of $k(\x,\x')$.
    \begin{align*}
        \sigma^2(\x) &= k(\x,\x)-k(\x,X)\left(K+\sigma_n^2I\right)^{-1}k(X,\x)\\
                    & = \bvphi^*(\x)\bvphi(\x)-\bvphi^*(\x)\phi_X\left(\phi_X^T\phi_X+\sigma_n^2I\right)^{-1}\phi^T_X\bvphi(\x)\\
                    & = \bvphi^*(\x)\bvphi(\x) - \bvphi^*(\x)\left(\phi_X\phi_X^T+\sigma_n^2I\right)^{-1}\phi_X\phi^T_X\bvphi(\x)\\
                    & = \sigma_n^2 \bvphi^*(\x)\left(\phi_X\phi_X^T+\sigma_n^2I\right)^{-1}\bvphi(\x),
    \end{align*}
    where $\phi_X = [\bvphi(\x_1),\bvphi(\x_2),\dots,\bvphi(\x_N)]$.
    Applying the product rule on the derivative of $\sigma^2(\x)$, leads to 
    \begin{align*}
        \frac{\diff\sigma^2(\x)}{\diff\x} &= \sigma_n^2\frac{\partial\bvphi^*(\x)}{\partial \x}\left(\phi^T_X\phi_X+\sigma_n^2I\right)^{-1} \bvphi(\x)\\
        &+ \sigma_n^2\bvphi^*(\x)\left(\phi^T_X\phi_X+\sigma_n^2I\right)^{-1} \frac{\partial\bvphi(\x)}{\partial \x}\\
        &\leq \frac{\partial\bvphi^*(\x)}{\partial \x}\bvphi(\x) + \bvphi^*(\x)\frac{\partial\bvphi(\x)}{\partial \x} \leq 2L_k,
    \end{align*}
    and the result follows.

    $\bm{(iii)}$
    This proof follows the one presented by \cite{Lederer2019}.
    From \cite{Snirivas2010}, we know that
    \begin{align*}
        \mathbb{P}\left\{|f([\x])-\mu([\x])|\leq 
    \beta_b^\frac{1}{2}\sigma([\x]), 
    ~\forall[\x]\in\mathcal{I}\right\}\geq 1-\delta.
    \end{align*}
    In addition, $f([\x])-L_f\tau\leq f(\bm{a})\leq f([\x])+L_f\tau,\, \forall \bm{a} \in [\x]$, $\mu(\bm{a})$ and $\sigma(\bm{a})$ analogously.
    Hence, for the left side of the inequality
    \begin{align*}
     &|f([\x])-\mu([\x])| + L_f\tau + \omega_\mu(\tau)\\
     \geq &|f([\x])+ L_f\tau-\mu([\x])-\omega_\mu(\tau)|\\ 
     \geq &|f(\x)-\mu(\x)|,
    \end{align*}
    and for the right side $\sigma([\x])-\omega_\sigma(\tau)+\omega_\sigma(\tau)\leq \sigma(\x)+\omega_\sigma(\tau)$.
    From this the result follows.
\end{IEEEproof}

In Proposition~\ref{proposition:Multi-task_uniform_error_bound}, proof of inequality $\bm{(ii)}$ shows that the influence of the data on $\omega_\sigma$ is insignificant if the observed data set is sparse in $\mathcal{X}$ and $\mathbb{N}_{\hil}$ is large, because this ensures that the rank of $\phi_X\phi_X^T$ is small (note that $\phi_X\phi_X^T$ is $N_\hil$ dimensional). This means that the first inequality is close to an equality.

The difference between the bounds in Proposition~\ref{proposition:Multi-task_uniform_error_bound} and \cite{Lederer2019} Theorem~3.1 lies in the definition of the moduli of continuity $\omega_{\mu}$ and $\omega_{\sigma}$ which are significantly reduced. Furthermore, these revised results simplify the provision of robust scaling factors for the multi-task setting in the following subsection.

\subsubsection*{Multi-Task}
To tackle the lack of knowledge of the true correlation matrix, a Bayesian approach is used similar to \cite{Capone2022}. Hence, Assumption~\ref{assum:continuity} is extended as follows:
\begin{assum}
\label{assum:sample_of_GP}
    The unknown vector valued function $\f:\mathcal{X}\to\mathbb{R}^u$ is a sample from a \gls{gp} with zero mean, multi-task kernel $K(\x,\x') = \Sigma k(\x,\x')$ and hyper-prior $\Sigma\sim p(\Sigma)$ with compact support, \ie, $\f(\x)\sim\mathcal{GP}(\bm{0},K(\x,\x'))$. The compact vector space $\mathcal{X}$ is equipped with a norm $\|\cdot\|_p,\, p\in\mathbb{N}$, and $\mathbb{R}^u$ is equipped with norm  $\|\cdot\|_q,\, q\in\mathbb{N}$. In addition, it is assumed that the base kernel $k(\x,\x')$ is at least four times partial differentiable on $\mathcal{X}$.
\end{assum}
The hyper-prior $p(\Sigma)$ encodes prior knowledge about the correlation between the tasks. Then, by using Bayesian inference, the hyper-posterior can be determined which allows for the construction of a confidence set $\mathcal{C}_\rho$ that comprises the true correlation matrix with probability $1-\rho$. Further details on the construction of $\mathcal{C}_\rho$ can be found at the end of this section.

Recall that, as in \cref{subsec:frequentist_statistics}, the goal is to investigate the influence of the correlation matrix on the scaling factor $\beta_b(\Sigma)$. Observe from Proposition~\ref{proposition:Multi-task_uniform_error_bound} that the moduli of continuity as well as the Lipschitz constant of the sample need to be adapted. The moduli of continuity both depend on the Lipschitz constant of the kernel, which is given by $L_K = qL_k$ where $q =\max_{i=1,\dots,u}(\Sigma_{ii})$ is the largest diagonal entry.
Hence,
\[ \omega_{\mu}(\tau) \leq \max_{\Sigma'\in\mathcal{C}_\rho} \sqrt{2\tau qL_k}\|\bmu_{\Sigma'}^{\Sigma'}\|_{\bhil_{\Sigma'}},\] and
\[\omega_{\sigma}(\tau) \leq \max_{\Sigma'\in\mathcal{C}_\rho}\sqrt{2\tau q L_k}.\] The maximization over the confidence set $\Sigma'\in\mathcal{C}_\rho$ ensures that the functions upper bound all moduli of continuty resulting from correlation matrices in $\mathcal{C}_\rho$.  
The last ingredient is the Lipschitz constant of the sample function $L_f$. In \cite{grunewalder10a,Lederer2019} the bound is obtained by first deriving the expected supremum of a sample from the differential kernel. This is achieved by making use of the metric entropy for the sample continuity \cite{Dudley1967}. After that, the Borell-TIS inequality \cite{Talagrand1994} is applied to bound the supremum from its expected value. In order to obtain similar results for vector valued functions from an \gls{icm} the procedure is slightly extended.

\begin{lemma}
\label{lemma:lipschitz_f}
Let Assumption~\ref{assum:sample_of_GP} hold, and let $\f: \mathcal{X} \to \mathbb{R}^u$, and recall that every function is a linear combination of single-task feature samples, \ie, $\f(\x) =  \sum_{i=1}^u \bm{b}_i h_i(\x)$, where $\bm{b}_i$ is the i\ts{th} column of the matrix $\Sigma^{\frac{1}{2}}$, $h_i(\x)\sim\mathcal{GP}(\bm{0},k(\x,\x'))$ are i.i.d.,  
and $L_h$ denotes the Lipschitz constant of the feature samples with probability $1-\delta$.
Then, we have with probability $1-\delta$ that
\begin{alignat*}{5}
    L_f \leq \max_{\Sigma\in\mathcal{C}_\rho}\left\| \Sigma^{\frac{1}{2}}\mathds{1} \right\|_{q} L_h, &\\
\end{alignat*}
where $\mathds{1}$ is a vector of ones, is an upper bound on the Lipschitz constants of all sample functions with correlation matrices from $\mathcal{C}_\rho$.
\end{lemma}

\begin{IEEEproof}
\begin{equation}
\begin{aligned}
        \max_{\x\in\mathcal{X}}\left\|\frac{\partial\f(\x)}{\partial \x}\right\|_{q} 
        &=\max_{\x\in\mathcal{X}}\left\|\sum_{i=1}^{u} \bm{b}_i \frac{\partial h(\x)}{\partial \x}\right\|_{q} \\
        &\leq\left\|\sum_{i=1}^{u} \bm{b}_i  \max_{\x\in\mathcal{X}}\left|\frac{\partial h(\x)}{\partial \x}\right|\right\|_{q} \\
        &\leq \left\|\Sigma^{\frac{1}{2}}\mathds{1} L_h\right\|_{q}\\
        &= \left\| \Sigma^{\frac{1}{2}}\mathds{1} \right\|_{q} L_h.
\end{aligned}
\label{eq:bound_sample_sup}
\end{equation}
Taking the maximum over $\mathcal{C}_\rho$ gives the result.
\end{IEEEproof}
Note that the Lipschitz constant of the feature processes $L_h$ can be derived with the methods used in \cite{grunewalder10a} or approximated using samples.
In Lemma~\ref{lemma:lipschitz_f} the dependency of the scaling parameter $\beta_b(\Sigma)$ on the correlation matrix was derived. The importance of the norms $p$ and $q$ will be discussed in \cref{subsec:practical_considerations}. 

Now we are able to state the main result of this section
\begin{theorem}
    \label{th:robust_scaling_factor}
    Under Assumption~\ref{assum:sample_of_GP}, $\omega_\mu,\omega_\sigma$ as defined before and $L_f$ as in Lemma~\ref{lemma:lipschitz_f}. Pick any $\Sigma'\in\mathcal{C}_\rho$ which should be used for inference and let 
    \begin{alignat*}{2}
        &\beta_b=2\log\left(\frac{|\mathcal{I}|}{\delta}\right) \qquad    &&\psi=L_f\tau+\omega_\mu(\tau)+\beta^\frac{1}{2}\omega_{\sigma}(\tau)\\
        &\nu^2: \text{ Lemma~\ref{lemma:nu}}\qquad
        &&\gamma^2 = \max_{\Sigma\in\mathcal{C}_\rho}\|\Sigma'^{-1}\Sigma\|_2
    \end{alignat*}
    Then, with $\bar{\beta}_b = (\nu+\gamma\beta_b^\frac{1}{2})^2$ we have with probability $(1-\delta)(1-\rho)$
    \begin{align*}
        |\f(\x)-\bmu^{\Sigma'}(\x)|\leq \bar{\beta}_b^\frac{1}{2}\bsig^{\Sigma'}(\x)+\psi, 
        \end{align*}
        for all $\x\in\mathcal{X}$.
    \end{theorem}
The proof of the theorem requires an additional lemma to compensate the uncertainty of the posterior mean.
    \begin{lemma}
        \label{lemma:nu}
        Let $\bmu^{\Sigma}_{\Sigma}\in \bhil_{\Sigma}, \bmu^{\Sigma'}_{\Sigma'}\in \bhil_{\Sigma'}$ be the expansion parameters of posterior means induced by the kernels $\Sigma k(\x,\x')$ and $\Sigma' k(\x,\x')$, respectively. Then, with $\nu^2 = \max_{\Sigma\in\mathcal{C}_\rho}\|\bmu^{\Sigma'}_{\Sigma'} - L\bmu^{\Sigma}_\Sigma\|_{\bhil_{\Sigma'}}^2 + \frac{1}{\sigma_n^2} \sum_{n=1}^N \|\bmu^{\Sigma'}(\x_n)-\bmu^{\Sigma}(\x_n)\|_2^2$, with $L$ as in (Lemma~\ref{lemma:linear_oper}), we have
        \begin{align*}
         |\bmu^{\Sigma'}(\x)-\bmu^{\Sigma}(\x)| \leq \nu\bsig^{\Sigma'}(\x).
        \end{align*}
    \end{lemma}
    \begin{IEEEproof}
        Using the reproducing property, we know
        \begin{align*}
            |\bmu^{\Sigma'}(\x)-\bmu^{\Sigma}(\x)|  \leq \bsig^{\Sigma'}(\x)\|\bmu^{\Sigma'}_{\Sigma',P}-L^P\bmu^{\Sigma}_{\Sigma,P}\|_{\bhil_{\Sigma'}^P}, 
        \end{align*}
        where ${\bhil_{\Sigma'}^P}$ is the \gls{rkhs} of the posterior kernel and $L^P :\bhil_{\Sigma}^P\to \bhil_{\Sigma'}^P := \Psi_{\Sigma'}^{\frac{1}{2}} L (\Phi_{\Sigma}\Phi_{\Sigma}^T+\sigma_n^2I)^{-\frac{1}{2}}$ with $\Psi_{\Sigma'} = (\Phi_{\Sigma'}\Phi_{\Sigma'}^T+\sigma_n^2I)$ is a bounded linear operator that corresponds to the posterior version of $L$ in Lemma~\ref{lemma:linear_oper}. The additional subscript $P$ in $\bmu^{\Sigma}_{\Sigma,P}$ denotes that the function $\bmu^{\Sigma}$ is reproduced by applying the evaluation functional of the posterior kernel.
        Using the linearity of inner products, we have
        \begin{align*}
            \|\bmu^{\Sigma'}_{\Sigma',P}-L^P\bmu^{\Sigma}_{\Sigma,P}\|_{\bhil_{\Sigma'}^P}^2 &= 
            \underbrace{\|\bmu^{\Sigma'}_{\Sigma',P}\|^2_{\bhil_{\Sigma'}^P}}_{(1)} + 
            \underbrace{\|L^P\bmu^{\Sigma}_{\Sigma,P}\|_{\bhil_{\Sigma'}^P}^2}_{(2)}\\ &- 
            \underbrace{2\langle \bmu^{\Sigma'}_{\Sigma',P},L^P\bmu^{\Sigma}_{\Sigma,P}\rangle_{\bhil_{\Sigma'}^P}}_{(3)}.
        \end{align*}
        For (1) we use from \cite{Snirivas2010} that
        \begin{align*}
            \|\bmu^{\Sigma'}_{\Sigma',P}\|^2_{\bhil_{\Sigma'}^P} &\leq \|\bmu^{\Sigma'}_{\Sigma'}\|_{\bhil_{\Sigma'}}^2 +\sum_{n=1}^N \|\bmu^{\Sigma'}(\x_n)\|_2^2.
        \end{align*}
        For (2) the term can be expanded as 
        \begin{align*}
            \|L^P\bmu^{\Sigma}_{\Sigma,P}\|_{\bhil_{\Sigma'}^P}^2 &= \langle L^P\bmu^{\Sigma}_{\Sigma,P},L^P\bmu^{\Sigma}_{\Sigma,P}\rangle_{\bhil_{\Sigma'}^P}\\
            &= \frac{1}{\sigma_n^2} \langle \Psi_{\Sigma'}^{\frac{1}{2}} L \bmu^{\Sigma}_{\Sigma},\Psi_{\Sigma'}^{\frac{1}{2}} L \bmu^{\Sigma}_{\Sigma}\rangle_{\bhil_{\Sigma'}^P}\\
            & = \frac{1}{\sigma_n^2} \langle L \bmu^{\Sigma}_{\Sigma}, (\Phi_{\Sigma'}\Phi_{\Sigma'}^T+\sigma_n^2I) L \bmu^{\Sigma}_{\Sigma}\rangle_{\bhil_{\Sigma'}^P}\\
            &= \|L \bmu^{\Sigma}_{\Sigma}\|_{\bhil_{\Sigma'}}^2+ \frac{1}{\sigma_n^2} \langle \bmu^{\Sigma}_{\Sigma},L^*\Phi_{\Sigma'}\Phi_{\Sigma'}^T L \bmu^{\Sigma}_{\Sigma}\rangle_{\bhil_{\Sigma'}}\\ 
            &= \|L \bmu^{\Sigma}_{\Sigma}\|_{\bhil_{\Sigma'}}^2+ \frac{1}{\sigma_n^2}\sum_{i=1}^n \|\bmu^{\Sigma}(\x_i)\|_2^2.
        \end{align*}
        The last term (3) can be expanded as 
        \begin{align*}
            &\langle \bmu^{\Sigma'}_{\Sigma',P},L^P\bmu^{\Sigma}_{\Sigma,P}\rangle_{\bhil_{\Sigma'}^P} = \frac{1}{\sigma_n}\langle \bmu^{\Sigma'}_{\Sigma',P},\Psi_{\Sigma'}^{\frac{1}{2}} L \bmu^{\Sigma}_{\Sigma}\rangle_{\bhil_{\Sigma'}^P}\\
            \quad &= \frac{1}{\sigma_n^2}\langle \Psi_{\Sigma'}^{-\frac{1}{2}}\Phi_{\Sigma'}\tilde{\y},\Psi_{\Sigma'}^{\frac{1}{2}} L\bmu^{\Sigma}_{\Sigma}\rangle_{\bhil_{\Sigma'}}\\
            \quad &= \frac{1}{\sigma_n^2}\langle \Phi_{\Sigma'}\tilde{\y},L\bmu^{\Sigma}_{\Sigma}\rangle_{\bhil_{\Sigma'}}\\
            \quad &= \frac{1}{\sigma_n^2}\langle \Psi_{\Sigma'}^{-1}\Phi_{\Sigma'}\tilde{\y},\Psi_{\Sigma'}L\bmu^{\Sigma}_{\Sigma}\rangle_{\bhil_{\Sigma'}}\\
            \quad &= \frac{1}{\sigma_n^2}\langle \bmu^{\Sigma'}_{\Sigma'},(\Phi_{\Sigma'}\Phi_{\Sigma'}^T+\sigma_n^2I) L\bmu^{\Sigma}_{\Sigma}\rangle_{\bhil_{\Sigma'}}\\
            \quad &= \langle \bmu^{\Sigma'}_{\Sigma'},L \bmu^{\Sigma}_{\Sigma}\rangle_{\bhil_{\Sigma'}} + \frac{1}{\sigma_n^2}\sum_{i=1}^n |\bmu^{\Sigma}(\x_i)^T\bmu^{\Sigma'}(\x_i)|^2. 
        \end{align*}
        Finally, we observe that the sum of all terms admit the quadratic equation given in the claim.
    \end{IEEEproof}
\begin{IEEEproof}
    Pick any $\Sigma'\in\mathcal{C}_\rho$ which is used for inference. The first step is to bound the posterior mean $\bmu^{\Sigma'}(\x)$ with respect to all members in $\mathcal{C}_\rho$, \ie, $|\f(\x)-\bmu^{\Sigma}(\x)|\leq |\f(\x)-\bmu^{\Sigma'}(\x)|+|\bmu^{\Sigma'}(\x)-\bmu^{\Sigma}(\x)|$. The latter term can be bounded by $\nu\bsig^{\Sigma'}(\x)$, see Lemma~\ref{lemma:nu}. 
    Next, the posterior standard deviation is bounded $\bsig^{\Sigma}(\x)\leq \gamma \bsig^{\Sigma'}(\x)$, see \cite{Luebsen2024} Lemma~1. Finally, the robust scaling factor $\bar{\beta}_b$ is derived by combining Lemma~\ref{lemma:lipschitz_f} with the previous derivations for $\omega_\mu$ and $\omega_\sigma$. 
\end{IEEEproof}

Theorem~\ref{th:robust_scaling_factor} provides the necessary components for calculating a robust scaling factor for multi-task \glspl{gp} using Bayesian statistics. The Lipschitz constants can be precomputed, but the values of $\gamma$ and $\nu$ (which can be calculated analytically) need to be updated whenever $\mathcal{C}_\rho$ or $\Sigma'$ change. This leads to the final part of this section, wherein practical details are specified.  

\subsection*{Practical Considerations}
\label{subsec:practical_considerations}
The importance of the norms defined on $\mathcal{X}$ lies in the resulting cardinality of the quotient set $\mathcal{I}$. The equivalence classes partition $\mathcal{X}$ into shapes depending on the norm, \eg, if $p=1$ the partitions would be rhombuses, if $p=2$ circles and if $p=\infty$ squares. Obviously, the cardinality changes with the norms. The minimal number of partitions to cover the input space is given by the covering number $N(\tau,\mathcal{X},\|\cdot\|_p)$.
For instance, if $p=\infty$ and $\mathcal{X}$ being a unit hypercube (which can always be obtained via input transformation) the covering number is given by $N(\tau,\mathcal{X},\|\cdot\|_\infty) = (\lceil 1/(2\tau) +1\rceil)^d$. 

The main advantage of the Bayesian framework is its ability to estimate $\mathcal{C}_\rho$ online by means of training data. Using the prior in Assumption~\ref{assum:continuity} and the likelihood $p(\tilde{\y}|\tX,\Sigma)$, a hyper-posterior $p(\Sigma|\tilde{\y},\tX)$ is obtained by applying Bayes rule
\begin{align*}
    p(\Sigma|\tX,\tilde{\y}) = \frac{p(\tilde{\y}|\tX,\Sigma) p(\Sigma)}{p(\tilde{\y}|\tX)},
\end{align*}
where $p(\tilde{\y}|\tX)$ is referred to the marginal likelihood. Bayes rule allows incorporating knowledge in form of data into the guess of the distribution of $\Sigma$. Usually, the hyperparameters of the \gls{map} are selected, which corresponds to the mode of $ p(\Sigma|\tX,\tilde{\y})$. Since the mode has measure zero, using the \gls{map} is not robust. For instance, if the distribution has slowly decaying tails, the error by only considering the mode increases dramatically. This encourages the use of the hyper-posterior $p(\Sigma|\tX,\tilde{\y})$ to compute a confidence set depending on the failure probability $\rho$. Note that the computation of $p(\Sigma|\tX,\tilde{\y})$ is typically challenging because the likelihood $p(\tilde{\y}|\tX,\Sigma)$ is Gaussian, while the prior $p(\Sigma)$ is not. Hence, the hyper-posterior can not be computed analytically, but approximated, \eg, by an empirical distribution, which is obtained by drawing samples using \gls{mcmc} methods.
From the empirical distribution, the confidence set $\mathcal{C}_\rho$ can be obtained by selecting the $\rho$-quantile. 

Many calculations require the maximization of eigenvalues over $\mathcal{C}_\rho$. As we have seen, the set is finite and consists in the numerical examples in \cref{sec:numerical_eval} of $\approx100$ samples. In addition, by using normalized $2\times 2$ matrices (this means that an additional task is considered, and the signal variances are equal which is reasonable in practice), the eigenspaces of all matrices are equal, which speeds up the computation significantly. The primary cause of the algorithm's slowdown is the presence of huge Gram matrices, that must be inverted during inference. This issue arises when the size of the data set $\bm{\mathcal{D}}$ is large, which occurs due to the incorporation of data from additional tasks. In future this can be mitigated by employing a sparse approximation of the Gram matrix, which is a common technique in \gls{gp} regression.

\section{Numerical Evaluation}
\label{sec:numerical_eval}

\subsection{Benchmark}
\begin{figure}[t]
    \centering
    \includegraphics{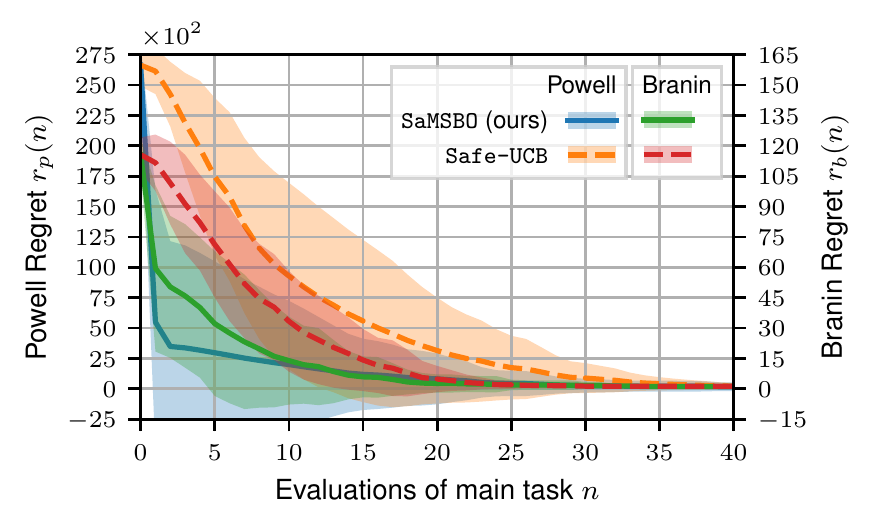}
    \caption{Comparison of the proposed multi-task safe \gls{bo} algorithm \texttt{SaMSBO} with the single-task equivalence \texttt{Safe-UCB} \cite{Sui2015} on the Powell and Branin function. The abscissa denotes the number of evaluations of the main task and the ordinates the best observation. The shaded area represents the standard deviation.}
    \label{fig:synth_fun_comparison}
\end{figure}

In this subsection, the proposed multi-task safe \gls{bo} algorithm is compared with its single-task equivalence. The benchmark is carried out on two synthetic but widely used functions, namely Powell and Branin. In the second part of this subsection, the algorithm is applied on a control problem, where PI controller parameters are safely tuned. The safe set is always defined as
\begin{align}
    \label{eq:safe_set}
    \mathcal{S}:= \{\x\in\mathcal{X}|\mu_1^{\Sigma'}(\x)+\bar{\beta}_b\sigma_1^{\Sigma'}(\x)\leq T\},
\end{align}
which means that no inputs should be evaluated where the corresponding function value exceeds the safety threshold $T$. We restrict safeness here only to the main task, \ie, the supplementary tasks are free to choose any input from $\mathcal{X}$. The dependency of the objective and the constraint is known, thus, one \gls{gp} suffices to model both functions. However, note that all statements in this work also apply if the dependency is unknown. 

Recall that the motivation is to reduce the evaluations of the main task due to high costs by incorporating information from supplementary tasks. As first order models always approximate the true system, we disturb the supplementary tasks. The general setting is for both tests the same: All data is at every iteration normalized to the unit interval and standardized to zero mean and unit variance at every iteration. Data normalization and standardization improve the performance and the numerical stability of the algorithm significantly. The base kernel utilized in the \gls{gp} is a squared exponential kernel. The prior over the correlation matrix is a \gls{lkj} distribution with a shape factor$\eta=0.1$ indicating that relatively high correlation is expected. The posterior samples are generated using the No-U-Turn Sampler algorithm \cite{Hoffman2014}, and $\bar{\beta}_b$ is computed in accordance with Theorem~\ref{th:robust_scaling_factor}. The failure probability of the confidence set is $\rho=0.15$, the failure probability $\delta=0.05$, the discretization $\tau=0.001$ and $\psi$ is neglected (due to dense discretization). The algorithm was executed for 40 iterations of the main task with 15 repetitions conducted with different starting inputs within the feasible region and randomly reinitialized disturbances. A single supplementary task is employed, with $2d$ evaluations per iteration. All implementations are carried out using GPyTorch \cite{gardner2018gpytorch} and BoTorch \cite{balandat2020botorch} frameworks in Python.


\begin{figure*}[t]
    \centering
    \NewDocumentCommand{\pll}{m}{
    \IfEq{#1}{1}{
        \node (bs#1) {};
    }{
        \node[right=0.1 of p#1] (bs#1) {};
    }
    \node[draw,circle,minimum size=0.2,right=0.5 of bs#1] (s#1) {};
    \node [draw,minimum width=1,minimum height=0.6,right=1. of s#1]  (C#1) {$C_{#1}(s)$};
    \node[draw,minimum width=1,minimum height=0.4,right=.5 of C#1] (G#1) {$G_{#1}(s)$};
    \node[draw,circle,minimum size=0.2,right=.5 of G#1] (so#1) {};
    \node[draw,minimum width=.6,minimum height=0.4, above=.5 of so#1] (F#1) {$F_{#1}(s)$};
    \node[right=0.5 of so#1] (out#1) {};
    
    \draw[-stealth] (bs#1.center) -- (s#1.west); 
    \draw[-stealth] (s#1.east) -- (C#1.west) node[pos=0.3,above] {$e_{#1}(t)$};
    \draw[-stealth] (C#1.east) -- (G#1.west);
    \draw[-stealth] (G#1.east) -- (so#1.west);
    \draw[-stealth] (F#1.south) -- (so#1.north);
    \draw[-stealth] (F#1.north) ++ (0,.5) -- (F#1.north) node[pos=0.1,above] {$w_{#1}(t)$};
    \draw[-stealth] (so#1.east) -- (out#1.east);
    \draw (out#1.center) ++(-0.3,0) -- ++(0,-.5) coordinate (t#1);
    \draw[-stealth] (t#1.north) -| (s#1.south) node[pos=0.95,right] {$-$};
}

\begin{tikzpicture}
    \pll{1};
    \node[right=0.1 of out1] (pN) {$\cdots$};
    \pll{N};
    \node[draw,circle,minimum size=0.2,right=0 of outN] (sum) {};
    \node at ($(sum.north west) + (-0.2,0.05)$) {$-$};

    \draw (bs1.center) ++(0.2,0) -- ++(0,-1) coordinate (temp);
    \draw[-stealth] (temp.center) -| (sum.south);
    \draw[-stealth] (sum.east) -- ++ (1,0) node[pos=0.9,above] {$z(t)$};
    \node[draw,minimum width=1,minimum height=0.6,left=0 of bs1.center] (Fr) {$F_r(s)$};
    \draw[-stealth] (Fr.west) ++ (-1.2,0) -- (Fr.west) node[pos=0.1, above] {$w_{N+1}(t)$};
    \node[draw, dashed, fit=(Fr)(sum), inner xsep=0.25cm, inner ysep=1.3cm, label={[anchor=north west] north west:{\Large$G_{cl}(s)$}}] (box) {};
\end{tikzpicture}
    \caption{Illustration of the interconnected system. The blocks $F_r$ and $F_i, i=1,\dots,N$ denote disturbance filters which colorize the white noise inputs $w_j, j=1,\dots,N+1$. $G_i$ denote the laser plants and $K_i$ PI controllers for each subsystem.}
    \label{fig:lbsync}
\end{figure*}
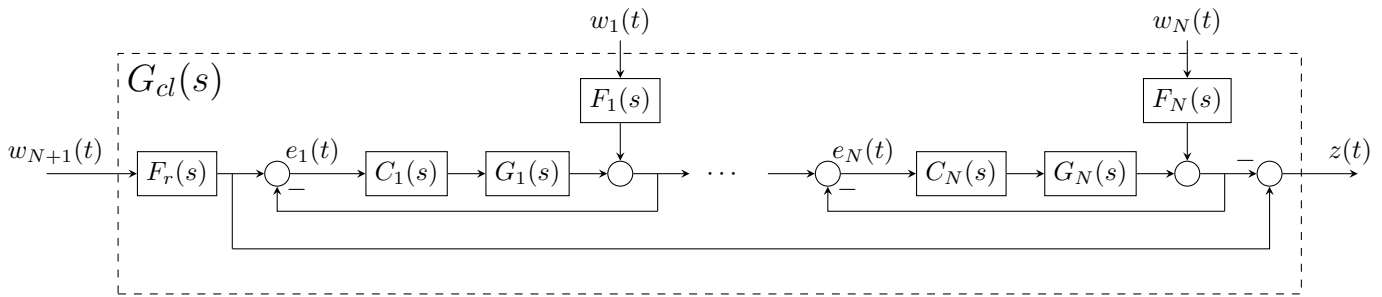

\subsubsection*{Synthetic Functions}
The Powell function is $d$-dimensional, with a global minimum at $x^*=(0,\dots,0)$. The Branin function is two-dimensional with three global minima. The input spaces are $\mathcal{X}=[-4,5]^d$ and $\mathcal{X}=[-5,10]\times[0,15]$, respectively. The safety thresholds are set to $T_\mathrm{P}=35.000$ and $T_\mathrm{B}=150$, respectively.  For the supplementary task, the functions are shifted in a random direction by a given disturbance factor, which denotes the relative shift with respect to the input space of the first task, \eg, for the Powell function a factor of $0.1$ means that the total shift amounts to $0.1(5-(-4))/2=\pm 0.45$.

\Cref{fig:synth_fun_comparison} summarizes the benchmark results. The median (dashed and solid lines) and the $90\%-10\%$ quantiles (shaded area) of the best observation over the number of main task evaluations is plotted. The proposed algorithm \textit{\textbf{s}afe \textbf{m}ulti-\textbf{s}ource \textbf{B}ayesian \textbf{o}ptimization} (\texttt{SaMSBO}) outperforms the single-task equivalent \texttt{Safe-UCB} in terms of solution quality and sample efficiency. The disturbance factor for the supplementary tasks is set to $0.3$ for both functions, indicating that the supplementary task is shifted significantly. Nevertheless, the results demonstrate that the proposed algorithm is still capable of reducing the number of evaluations of the main task. For smaller disturbances, this effect would be even higher \cite{Luebsen2024}.

\subsubsection*{Laser-Based Synchronization at European X-Ray Free Electron Laser}
In this section the algorithms are applied on a control problem. The goal is to optimize the synchronization of a chain of lasers which are exposed to disturbances similar to the \gls{lbsync} at \gls{euxfel}. The considered plant is depicted in \cref{fig:lbsync} where $G_i$ represent the laser models and $K_i$ denote PI controllers. The number of subsystems is set to $N=5$, which implies a ten-dimensional optimization problem. The filter models $F_r$ and $F_{1:N}$ colorize the white Gaussian noise inputs $w_{1:N+1}$ to model environmental disturbances, e.g., vibrations, temperature changes and humidity. In order to mimic the discrepancy between simulation and reality, the primary task employs nominal models, while supplementary tasks are subjected to disturbances. It is assumed that uncertainty resides in the filter models, given that the laser model can be accurately identified and exhibits minimal variation over time, while disturbance sources change more frequently. The disturbance acts on the system matrices of the filter models and is randomly initialized, where the disturbance factor denotes the percentage of the system matrices that are added/subtracted. The safety threshold is set to $T_L=40$. The goal is to minimize the root-mean-square seminorm of the performance output $z$ by tuning the PI parameters of the controllers $K_{1:N}$. Following Parseval's Theorem this corresponds to an $H_2$ minimization of the closed-loop system $G_{cl}(s)$ \cite{Heuer2018}.

\begin{figure}[t]
    \centering
    \includegraphics{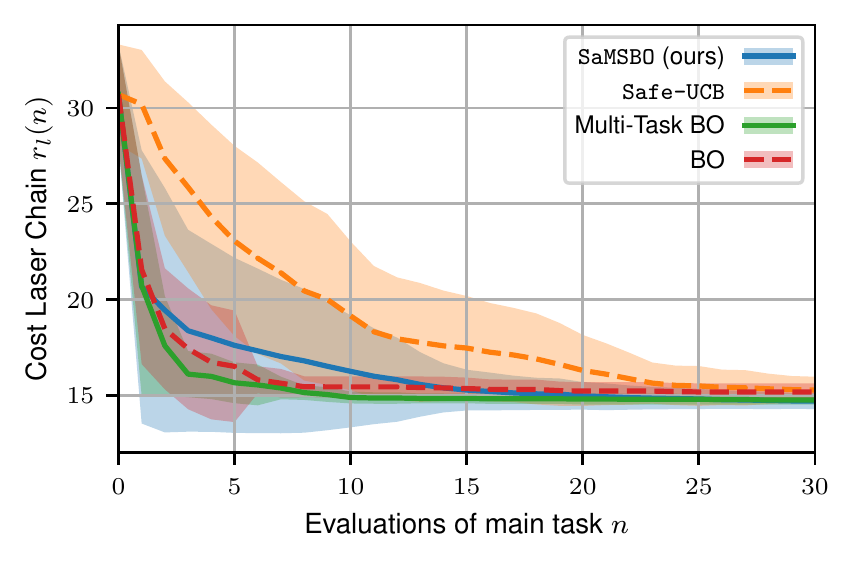}
    \caption{Comparison of the proposed multi-task safe \gls{bo} algorithm \texttt{SaMSBO} with its single-task equivalent \texttt{Safe-UCB} \cite{Sui2015} and none-safe single- and multi-task \gls{bo} algorithms. The multi-task \gls{bo} algorithms have access to one supplementary task which is disturbed by 0.3. The abscissa denotes the number of evaluations of the main task and the ordinate the best cost value. The lines and dashed lines denote the average and the shaded area represents the standard deviation.}
    \label{fig:comp_laser_chain}
\end{figure}

The laser chain benchmark is presented in \Cref{fig:comp_laser_chain}, which also compares both safe and non-safe single- and multi-task equivalences. Again, the algorithms are 15 times applied with different initial inputs and reinitialized disturbances. The multi-task \gls{bo} algorithms have access to one supplementary task which is disturbed by $0.3$. By surprise, \texttt{SaMSBO} even outperformed its none safe equivalent. This can be explained that the main task was forced to exploit due to the safety constraints, whereas in multi-task \gls{bo} the main task performed exploration. It is worth noting that both safe \gls{bo} algorithms never violated the safety constraints, whereas the none safe violated it 300 (multi-task) and 343 (single-task) times.

One possible argument for the multi-tasks good performance is that the disturbances of the filters have no influence on the cost value. However, this is not the case; the cost values show significant discrepancies. The reason for the good performance is the shape of the cost function, which is relatively flat over a large region. Therefore, the supplementary task is used to identify the location of this area, and the multi-task \gls{gp} transfers this knowledge to the primary task and guide the expansion of the safe region in the direction of the optimal area. In other words, the supplementary task is used to identify the safe region in which the primary task is used to identify the optimal solution. Hence, the extra evaluations needed for exploration are reduced.  
Clearly, the knowledge transfer is limited, as only correlations between the tasks can be identified, which translates to affine transformations. Nevertheless, we demonstrated in this section that this approach is more robust than expected, as evidenced by the shift of the nonlinear functions Powell and Branin, which are clearly not affine transformations. It is still possible to extract information from the supplementary task. However, if the discrepancies exceed a certain degree, the performance of this approach will decrease which is a general limitation of the \gls{icm}.

\section{Conclusion and Outlook}
In this manuscript we proposed a novel multi-task safe \gls{bo} algorithm, \texttt{SaMSBO}, which is capable of safely optimizing a primary task while incorporating information from supplementary tasks. We investigated the influence of uncertain correlation matrices under the frequentists and Bayesian view.
Our theoretical derivations are underlined with numerical evaluations which showed that, in comparison to single-task optimization, the proposed algorithm is able to reduce the number of evaluations of the primary task, which makes it very suitable for optimizations where the evaluation of the primary task is expensive. 

In future work the analysis can be extended for more complex kernels, \eg, the linear model of co-regionalization. Moreover, the computation of the hyperposterior need to be accelerated which can be achieved by applying variational inference. Furthermore, the computational efficiency need to be improved by using sparse approximations. Finally, other methods for correlation detection can be investigated.

\bibliographystyle{IEEEtran}
\bibliography{IEEEabrv,biblio}

\appendix

\section{Frequentist Safety Bounds for Multi-Task Kernels}


\subsection{Proof of Lemma~\ref{lemma:frequentist_multi-task}}
\label{proof:lemma_frequentist_multi-task}
\begin{IEEEproof}
We consider the difference between of the posterior mean and the true function value for the $i\ts{th}$ task.
	\begin{align*}
		|f_i(\x)-\mu_i(\x)| &= |\bm{e}^T_i \bm{f}(\x)-\bm{e}^T_i\bm{\mu}(\x)|\\ 
		&= |\bm{e}^T_i\langle \Phi(\x),\bm{f}\rangle_{\bhil} -\bm{e}^T_i\langle \Phi(\x),\bm{\mu}\rangle_{\bhil}|\\
		 &= |\bm{e}^T_i\langle \Phi(\x), \bm{f}-\bm{\mu}\rangle_{\bhil}|\\ 
		 &= |\bm{e}^T_i\Phi^*(\x)(\bm{f}-\bm{\mu})|\\ 
		 &= |\bvphi_i^*(\x)(\bm{f}-\bm{\mu})|.
	\end{align*}
	Furthermore, we know that $\bm{\mu}(\x) = \Phi^*(\x)\Phi_{\tX}(\Phi_{\tX}^T\Phi_{\tX}+\sigma_n^2I)^{-1}\tilde{\bm{y}}$. With $\tilde{\bm{y}} = \Phi_{\tX}^T\bm{f}+\bm{\varepsilon}$ we have,
	\begin{align*}
		&|\bvphi_i^*(\x)\f-(\bvphi_i^*(\x)\Phi_{\tX}(\Phi_{\tX}^T\Phi_{\tX}+\sigma_n^2I)^{-1})(\Phi_{\tX}^T\f+\bm{\varepsilon})| \\
		\leq &\underbrace{|\bvphi_i^*(\x)(I-\Phi_{\tX}(\Phi_{\tX}^T\Phi_{\tX}+\sigma_n^2I)^{-1}\Phi_{\tX}^T)\bm{f}|}_{(1)}\\
		&+\underbrace{|\bvphi_i^*(\x)\Phi_{\tX}(\Phi_{\tX}^T\Phi_{\tX}+\sigma_n^2I)^{-1}\bm{\varepsilon}|}_{(2)}.
	\end{align*}

	For (1) similar to \cite{chowdhury17a} we have
	\begin{align*}
		&|\bvphi_i^*(\x)(I-\Phi_{\tX}(\Phi_{\tX}^T\Phi_{\tX}+\sigma_n^2I)^{-1}\Phi_{\tX}^T)\bm{f}|\\
		 = \, & |\bvphi_i^*(\x)(I-(\Phi_{\tX}\Phi_{\tX}^T+\sigma_n^2I)^{-1}\Phi_{\tX}\Phi_{\tX}^T)\bm{f}|\\
		 = \, & |\sigma_n^2\bvphi_i^*(\x)(\Phi_{\tX}\Phi_{\tX}^T+\sigma_n^2I)^{-1}\bm{f}|\\
		 \leq \, & \|(\Phi_{\tX}\Phi_{\tX}^T+\sigma_n^2I)^{-1}\bvphi_i(\x)\|_{\hil}\|\bm{f}\|_{\bhil}\\
		 \leq \, & \|\bm{f}\|_{\bhil}\sigma_i(\x).
	\end{align*}
	This proves the first term on the right side of the theorem.\\
	Furthermore, (2) can be upper bounded by
	\begin{align*}
		&|\bvphi_i^*(\x)\Phi_{\tX}(\Phi_{\tX}^T\Phi_{\tX}+\sigma_n^2I)^{-1}\bm{\varepsilon}|\\
		&\quad = |\bvphi_i^*(\x)(\Phi_{\tX}\Phi_{\tX}^T+\sigma_n^2I)^{-1}\Phi_{\tX}\bm{\varepsilon}|\\
		&\quad \leq  \|\bvphi_i^*(\x)(\Phi_{\tX}\Phi_{\tX}^T+\sigma_n^2I)^{-1}\Phi_{\tX}\|_\mathcal{H} \|\bm{\varepsilon}\|_2\\
		&\quad = (\bvphi_i^*(\x)(\Phi_{\tX}\Phi_{\tX}^T+\sigma_n^2I)^{-1}\Phi_{\tX}\Phi_{\tX}^T\\ &\qquad \qquad(\Phi_{\tX}\Phi_{\tX}^T+\sigma_n^2I)^{-1}\bvphi_i(\x))^{1/2} \|\bm{\varepsilon}\|_2\\
		&\quad \leq  (\bvphi_i^*(\x)(\Phi_{\tX}\Phi_{\tX}^T+\sigma_n^2I)^{-1}(\Phi_{\tX}\Phi_{\tX}^T+\sigma_n^2I)\\ &\qquad \qquad(\Phi_{\tX}\Phi_{\tX}^T+\sigma_n^2I)^{-1}\bvphi_i(\x))^{1/2} \|\bm{\varepsilon}\|_2\\
		&\quad =  (\bvphi_i^*(\x)(\Phi_{\tX}\Phi_{\tX}^T+\sigma_n^2I)^{-1}\bvphi_i(\x))^{1/2} \|\bm{\varepsilon}\|_2\\
		&\quad = \sigma_n^{-1} (\sigma_n^2\bvphi_i^*(\x)(\Phi_{\tX}\Phi_{\tX}^T+\sigma_n^2I)^{-1}\bvphi_i(\x))^{1/2} \|\bm{\varepsilon}\|_2\\
		&\quad = \sigma_n^{-1} \sigma_i(\x) \|\bm{\varepsilon}\|_2,
	\end{align*}
	where $\sigma_i(\x)$ is the posterior standard deviation.\\
	Since $\bm{\varepsilon}$ denotes a vector of random variables, we employ a concentration inequality \cite{Hsu2011} which gives us the following with probability at least $1-\delta$:
	\begin{align*}
		\|\bm{\varepsilon}\|_2^2\leq\sigma_n^2\left(N+2\sqrt{N}\sqrt{\ln{\frac{1}{\delta}}}+2\ln{\frac{1}{\delta}}\right).
	\end{align*}
	Putting all together and repeating for every $i$ the result follows.
\end{IEEEproof}

\section{Miscellaneaous}


\subsection{Auxiliary Results}
\begin{corollary}
 \label{corollary:equivalent_RKHS}
     Let $\bhil_{\Sigma'}$, $\bhil_{\Sigma}$ be two multi-task RKHS with kernels $K(\x,\x') = \Sigma'\otimes k(\x,\x')$ and $K(\x,\x') = \Sigma\otimes k(\x,\x')$, respectively with $\Sigma,\Sigma'\in\mathcal{L}_{+}(\mathbb{R}^n)$ being symmetric and positive definite. Then, 
     \begin{align*}
         \bhil_{\Sigma'} = \bhil_{\Sigma}.
     \end{align*}
 \end{corollary}
 \begin{IEEEproof}
     From \cite{Berlinet2004} we know that $\bhil_{\Sigma'}\subseteq \bhil_{\Sigma}$ is equivalent to the existence of a finite constant $\beta$, such that $\beta^2K_\Sigma(\x,\x')-K_{\Sigma'}(\x,\x')$ is a positive definite kernel. Clearly, this is satisfied by choosing $\beta^2\geq \|\Sigma'\Sigma^{-1}\|_2^2$. Since $\Sigma,\Sigma'$ are positive definite, this proves the first direction. Moreover, there exists with the same argumentation a finite constant $\alpha$ such that $\alpha^2 K_{\Sigma'}(\x,\x')-K_{\Sigma}(\x,\x')$ is a positive definite kernel which indicates that $\bhil_{\Sigma}\subseteq \bhil_{\Sigma'}$. Hence, we have $\bhil_{\Sigma'} = \bhil_{\Sigma}$.   
 \end{IEEEproof}

\end{document}